\newcommand\makeExtended[0]{1}
\documentclass[runningheads,dvipsnames]{llncs}

\newcommand{\contentBasePath}[0]{content/}

\usepackage[utf8]{inputenc} 
\usepackage[english]{babel}
\usepackage{graphicx} 
\usepackage{xcolor}
\usepackage{booktabs}
\usepackage{multirow}
\usepackage{multicol}
\usepackage{subfig} 
\usepackage{ifthen}
\usepackage{amsmath}

\newboolean{requirement_aaai}
\ifthenelse{\isundefined{\makeRequirementAAAI}} 
	{\setboolean{requirement_aaai}{false}}
	{\setboolean{requirement_aaai}{true}}
	
\newboolean{requirement_kes}
\ifthenelse{\isundefined{\makeRequirementKES}} 
	{\setboolean{requirement_kes}{false}}
	{\setboolean{requirement_kes}{true}}

\newboolean{article_extended}
\ifthenelse{\isundefined{\makeExtended}}{
	\setboolean{article_extended}{false}}{
	\setboolean{article_extended}{true}}

\newboolean{requirement_twocolumn}
\ifthenelse{\isundefined{\makeRequirementTwocols}}{
	\setboolean{requirement_twocolumn}{false}}{
	\setboolean{requirement_twocolumn}{true}}

\ifthenelse{\boolean{requirement_aaai}}{
    \newcommand{\autoref}[1]{\ref{#1}}
    \newcommand{\href}[2]{\url{#1}}
}{
    \usepackage[bookmarks=false]{hyperref}
        \hypersetup{colorlinks,
          linkcolor=blue,
          citecolor=blue,
          urlcolor=blue}
}

\newboolean{vc_is_included}
\gdef\GITAbrHash{432623d}%
\gdef\VCDateTEX{2023/07/04}%
}{
	\setboolean{vc_is_included}{false}
	
	\gdef\GITAbrHash{}

	\gdef\VCDateTEX{}

}

\newcommand{\articleTitle}[0]{GRAN is superior to GraphRNN: node orderings, kernel- and graph embeddings-based metrics for graph generators}

\newcommand{\articleSubtitle}[0]{Kernel- and Graph Embeddings-based Metrics for Graph Generators}

\newcommand{\articleAuthorSpringer}[0]{Ousmane Touat\inst{2} \and Julian Stier\inst{1}\orcidID{0000-0001-5710-9240} \and Pierre-Edouard Portier\inst{2} \and Michael Granitzer\inst{1}\orcidID{0000-0003-3566-5507}}
\newcommand{\articleAuthorRunning}[0]{Touat et al.}
\newcommand{\articleEmail}[0]{ousmanetouat@outlook.com}
\newcommand{\articleInstituteSpringer}[0]{
University of Passau\\
\email{julian.stier@uni-passau.de}
\and INSA Lyon\\
\email{\articleEmail}
\ifthenelse{\boolean{vc_is_included}}{{\tiny\protect\\\tiny\VCDateTEX~$\sim$~\GITAbrHash}\vspace{-1em}}{~}
}
\newcommand{\articleRepo}[0]{https://github.com/otouat/GNNEvaluationMetrics/}

\makeatletter
\renewcommand{\boxed}[1]{\text{\fboxsep=.2em\fbox{\m@th$\displaystyle#1$}}}
\newcommand{\smallbigcirc}[1]{%
  \vcenter{\hbox{\scalebox{0.77778}{$\m@th#1\bigcirc$}}}%
}
\newcommand{\make@circled}[2]{%
  \ooalign{$\m@th#1\smallbigcirc{#1}$\cr\hidewidth$\m@th#1#2$\hidewidth\cr}%
}
\newcommand{\ostar}{\mathbin{\mathpalette\make@circled\star}}
\makeatother

\DeclareSymbolFont{extraup}{U}{zavm}{m}{n}
\DeclareMathSymbol{\varheart}{\mathalpha}{extraup}{86}
\DeclareMathSymbol{\vardiamond}{\mathalpha}{extraup}{87}

\definecolor{Green}{rgb}{0.01, 0.75, 0.24}
\definecolor{Red}{rgb}{0.76, 0.23, 0.13}
\definecolor{Gray}{rgb}{0.41, 0.41, 0.41}

\usepackage{amsmath,amsfonts,bm}

\def\eqref#1{equation~\ref{#1}}

\def\1{\bm{1}}

\DeclareMathAlphabet{\mathsfit}{\encodingdefault}{\sfdefault}{m}{sl}
\SetMathAlphabet{\mathsfit}{bold}{\encodingdefault}{\sfdefault}{bx}{n}

\usepackage[T1]{fontenc}
\begin{document}
\title{\articleTitle}
\titlerunning{\articleSubtitle}
%
\author{\articleAuthorSpringer}

%
\authorrunning{\articleAuthorRunning}
%
\institute{\articleInstituteSpringer}

%
\maketitle              
\begin{abstract}
A wide variety of generative models for graphs have been proposed. 
They are used in drug discovery, road networks, neural architecture search, and program synthesis.
Generating graphs has theoretical challenges, such as isomorphic representations -- evaluating how well a generative model performs is difficult.
Which model to choose depending on the application domain?

We extensively study \textit{kernel-based metrics} on distributions of graph invariants and \textit{manifold-based} and \textit{kernel-based metrics} in \textit{graph embedding space}.
Manifold-based metrics outperform kernel-based metrics in embedding space.
We use these metrics to compare GraphRNN and GRAN, two well-known generative models for graphs, and unveil the influence of node orderings.
It shows the superiority of GRAN over GraphRNN - further, our proposed adaptation of GraphRNN with a depth-first search ordering is effective for small-sized graphs.


A guideline on good practices regarding dataset selection and node feature initialization is provided. 
Our work is accompanied by open-source code and reproducible experiments.



\end{abstract}

\section{Introduction}\label{sec:introduction}
For a few years now, graph generation via deep generation models has been a subject of particular attention, notably because of its promising applications, especially in pharmacy. 
Beyond these specified applications, graphs are also central to many fields, such as physics or sociology. 

The attention on graph generation has been accentuated with the arrival of autoregressive generation models \cite{DBLP:journals/corr/abs-1802-08773,DBLP:journals/corr/abs-1910-00760,li2018learning} , able to build a graph with a sequence of elementary steps (e.g., the addition of a node and its connection to the current graph). 
These models are among the most scalable graph generator while still being able to approach graph distribution better. 
The main challenge of these generators is that since they must learn to generate graphs in sequence, these models deal with the problem of non-unique representation of graphs, where the node orderings dictate how to represent the graph in sequence. 
Since there can be many such orderings and, thus, ways of sequentially representing graphs, the sequential generation problem can quickly become challenging to model.
One approach is to model the problem by estimating a lower bound on the maximum likelihood, considering only a subset of the set of node orderings of a graph.

Another challenge is the evaluation of the performance of these models. 
Until now, the primary method of evaluation consists in using metrics based on the computation of kernels from statistical measures related to the studied graphs, computing, for example, the Maximum Mean Discrepancy of the node degree distribution \cite{DBLP:journals/corr/abs-1802-08773,DBLP:journals/corr/abs-1910-00760,goyal2020graphgen,obray2021evaluation}. 
With advances in graph neural networks (GNN), evaluation methods using data embeddings, originally applied to image generation models, have recently been transposed to the domain of generative models for graphs \cite{ggmiem,Thompson2020BuildingLU}. 
Nevertheless, these techniques still need to be sufficiently studied, and their use is just beginning to spread.

Our \textbf{contributions} comprise 1st/ extensive studies on recent evaluation techniques based on graph embeddings which are learned through pre-trained graph classifiers, on several factors such as the pre-training dataset or the node feature initialization technique, giving insights on good practices for graph evaluation accompanied with a \href{\articleRepo}{\textit{reproducible code repository}}. 2nd/ \textit{independent} and \textit{new empirical studies} on GraphRNN and GRAN compared using several node orderings, adding experiments with a \textit{newly proposed node ordering} (depth-first search) on GraphRNN, using both graph-embeddings- and kernel-based evaluation techniques.
These contributions show new insights on graph embeddings-based metrics for generative models of graphs and motivate using depth-first-search as a node ordering when learning GraphRNN on small graph datasets.

\section{Metrics for Generative Models of Graphs}\label{sec:graph-eval}
Generative models of graphs are trained on graph datasets and are then used to sample new graphs.
Good evaluation metrics should be able to rank graph generative models on how well they capture the characteristics of the underlying graph distribution $P_G$ from the training set samples $X^{train}_G\sim P_G$.
The graph edit distance is a standard similarity measure.
However, in general, its computation is NP-hard, \cite{zeng2009comparing}.
First papers on deep graph generation used graph statistics such as node degree and clustering coefficients to measure the performance of generative models \cite{bojchevski2018netgan}.

\subsection{Kernel-based Evaluation Metrics}
In their work, You et al. employ Maximum Mean Discrepancy \textbf{MMD} as a measure of distance between graph invariants, such as node degree or the number of 4-orbits \cite{DBLP:journals/corr/abs-1802-08773}. This allow them to assess how effectively the generated graphs from their trained model capture properties found in reference graphs.
MMD between two drawn distribution $x$ and $y$ based on a kernel $k$ is defined as:
\ifthenelse{\boolean{requirement_twocolumn}}{
    \begin{equation}
    \begin{aligned}
        \operatorname{MMD}^{2}(p \| q) = \mathbb{E}_{x, y \sim p}&[k(x, y)] + \mathbb{E}_{x, y \sim q}[k(x, y)] \\
         & - 2 \mathbb{E}_{x \sim p, y \sim q}[k(x, y)]
    \end{aligned}
    \end{equation}
}{
    \begin{equation}
        \operatorname{MMD}^{2}(p \| q) =\mathbb{E}_{x, y \sim p}[k(x, y)] + \mathbb{E}_{x, y \sim q}[k(x, y)] - 2 \mathbb{E}_{x \sim p, y \sim q}[k(x, y)]
    \end{equation}
}
For graphs, $p$ and $q$ refer to the distributions of computed graph invariants, such as the node degree distribution from the generated graphs and the set of reference graphs.
Those metrics are currently widely used in the graph generation community \cite{obray2021evaluation}.

\subsection{Graph Embeddings-based Evaluation Metrics}
We introduce the graph embeddings-based metrics used in our study:
\textit{Fréchet Distance}, \textit{Precision}, \textit{Recall}, \textit{Density}, \textit{Coverage} and \textit{F1-Score} use graph representations learned by the \textit{Graph Isomorphism Network} model.

\paragraph{Graph Isomorphism Network}
Graph Isomorphism Network (GIN) is a robust message-passing neural network architecture, usually designed for graph classification tasks \cite{xu2019powerful}.
We define $G=(V, E)$ as an undirected graph defined by its set of $n$ nodes denoted $V$, and a set of $e$ edges $E$.
A node $v_{i} \in V$ have its $d$-dimensional node feature $X \in \mathbb{R}^{e\times d}$.
The neighborhood $\mathcal{N}(v_i)$ is a set of nodes connected to $v_i$.
The hidden representation at the $k$-th layer $h^{(k)}(v_{i})$, with $\phi^{(k)}$ being the node aggregation function, is modelized using a multilayer perceptron (MLP):
\ifthenelse{\boolean{requirement_twocolumn}}{
    \begin{equation}\label{eqn:gin-hidden}
    \begin{aligned}
        h^{(k)}(v_{i}) = \text{MLP}^{(k)}(&(1+\epsilon^{(k)})h^{(k-1)}(v_{i}), \\
            & ~ \phi^{(k)}(\{h^{(k)}(u) ~|~ u \in \mathcal{N}(v_{i})\}))
    \end{aligned}
    \end{equation}
}{
    \begin{equation}\label{eqn:gin-hidden2}
        h^{(k)}(v_{i}) = \text{MLP}^{(k)}((1+\epsilon^{(k)})h^{(k-1)}(v_{i}), ~ \phi^{(k)}(\{h^{(k)}(u) ~|~ u \in \mathcal{N}(v_{i})\}))
    \end{equation}
}
The graph level readout $\xi_{rd}$, obtaining graph-level representations, ``can be a simple permutation invariant function such as summation or a more sophisticated graph-level pooling function'' \cite{xu2019powerful} that can be either a sum or an average over graph representations of each layer.
We get the graph's embeddings by performing such sum ``$\sum$'' or concatenation ``$\mathbin\Vert$'' over pooled node representations at each layer of the GIN \cite{xu2019powerful}:
\ifthenelse{\boolean{requirement_kes}}{
    \begin{minipage}{.48\linewidth}
    \begin{equation}\label{eqn:readout-concat}
        x_{\mathbin\Vert} = \boldsymbol{\mathbin\Vert}\left(\xi_{rd}(\{h^{(k)}(u) ~|~ u \in N(v_{i}) \}) ~|~ l \in \{1,2,..,K\}\right)
    \end{equation}
    \end{minipage}%
    \begin{minipage}{.5\linewidth}
    \begin{equation}\label{eqn:readout-sum}
        x_{\sum} = \sum_{l=1}^{K} \xi_{rd}(\{h^{(k)}(u) ~|~ u \in N(v_{i}) \})
    \end{equation}
    \end{minipage}
}{
    \begin{equation}\label{eqn:readout-concat}
        x_{\mathbin\Vert} = \boldsymbol{\mathbin\Vert}\left(\xi_{rd}(\{h^{(k)}(u) ~|~ u \in N(v_{i}) \}) ~|~ l \in \{1,2,..,K\}\right)
    \end{equation}
    \begin{equation}\label{eqn:readout-sum}
        x_{\sum} = \sum_{l=1}^{K} \xi_{rd}(\{h^{(k)}(u) ~|~ u \in N(v_{i}) \})
    \end{equation}
}
For the first expression, Xu et al. proves that GIN is equivalent to the Weisfeiler-Lehman Isomorphism Test (1-WL) \cite{weisfeiler_lehman}.

\paragraph{Pre-training GIN for Graph Feature Extraction}
To evaluate the quality of generative image models, metrics such as the \textit{Fréchet Inception Distance} (FID) \cite{heusel2018gans} have been used and are computed based on representations from image embedding space.
Image embeddings are obtained from learned image classification models such as Inception v3 on ImageNet \cite{DBLP:journals/corr/SzegedyVISW15}.
\ifthenelse{\boolean{requirement_twocolumn}}{\begin{figure*}[tbh]}{\begin{figure}[tbh]}
    \centerline{
        \ifthenelse{\boolean{requirement_twocolumn}}{
            \includegraphics[width=0.9\linewidth]{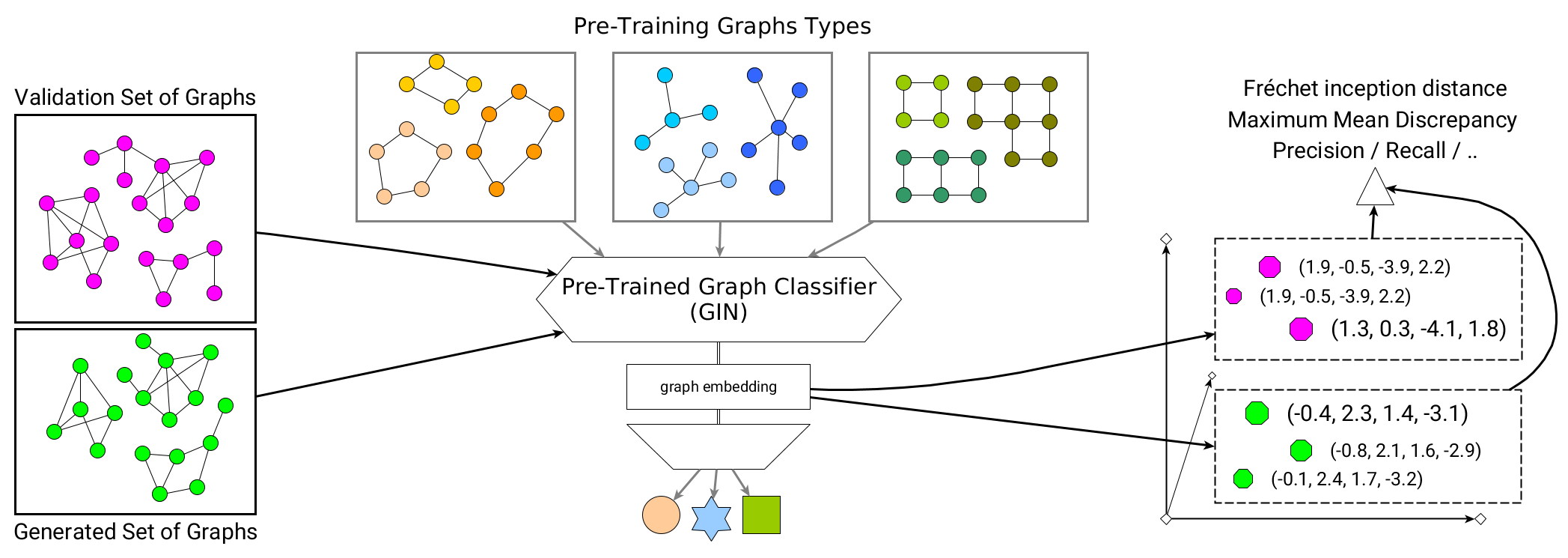}
        }{
            \includegraphics[width=0.9\linewidth]{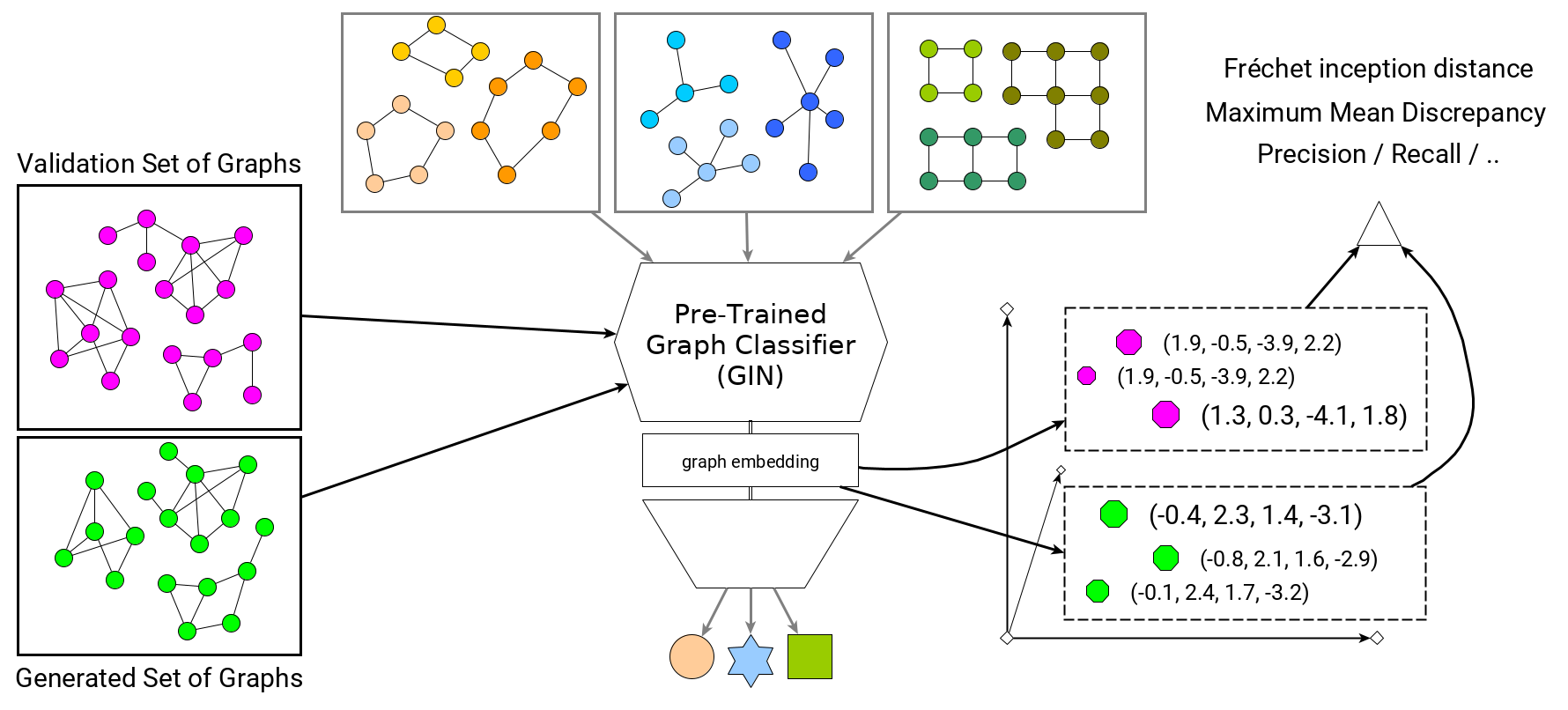}
        }
    }
    \caption{
    Embedding-based metrics are computed in embedding spaces of a GNN such as the GIN \cite{xu2019powerful}.
    After training this GNN on classifying types of graphs, we compare two sets of graphs by computing metrics based on their graph representations -- a $N\times d$ matrix, with $N$ the number of graphs and $d$ the graph manifold size.
    The reference set of graphs is compared to a trained model generated set of graphs, to assert the generative model performance.
    }
    \label{fig:gnnmetrics} 
    \vspace{-1.5em}
\ifthenelse{\boolean{requirement_twocolumn}}{\end{figure*}}{\end{figure}}
This approach was first explored on small graphs using GIN \cite{ggmiem}.
Similarly, we use metrics in \textit{graph} embedding space as sketched in figure \ref{fig:gnnmetrics}.
\textit{Graph} embeddings are extracted from a graph neural network trained on a multi-class classification task, where each class represents one type of graph, which has also been used in training for the generative tasks (i.e., grid graphs, community graphs).

\paragraph{Graph Embeddings-based Metrics}
Let $(X_1,\cdots, X_N)$ $\in\mathbb{R}^{N\times d}$ be vector representations of size $d$ in the graph embedding space of the pre-trained GIN obtained from graphs $(G_1, \cdots, G_N)$ through equations \ref{eqn:readout-concat} or \ref{eqn:readout-sum}.

The first studied metric, called Fréchet Distance \textit{FD}, is a popular metric derived from \textit{FID} \cite{heusel2018gans} used in the image generation domain.
This metric compares both generated and authentic samples of graphs in the graph embedding space, taken as Gaussian distributions with means and covariances $\boldsymbol{p}_{gen} = (m_{gen}, C_{gen})$ and $\boldsymbol{p}_{real} = (m_{real}, C_{real})$.
Fréchet Distance, accounting for the quality and diversity of generated samples, is computed as: 
\ifthenelse{\boolean{requirement_twocolumn}}{
    \begin{equation}\label{eqn:gin-hidden3}
    \begin{aligned}
        d^{2}\left(\boldsymbol{p}_{gen}, \boldsymbol{p}_{real}\right) = &\left\|{m}_{gen}-{m}_{real}\right\|_{2}^{2}+\operatorname{Tr}({C}_{gen}+\\
        & {C}_{real}-2\left({C}_{gen}{C}_{real}\right)^{1 / 2})
    \end{aligned}
    \end{equation}
}{
    \begin{equation}
        d^{2}\left(\boldsymbol{p}_{gen}, \boldsymbol{p}_{real}\right) = \left\|{m}_{gen}-{m}_{real}\right\|_{2}^{2}+\operatorname{Tr}\left({C}_{gen}+{C}_{real}-2\left({C}_{gen}{C}_{real}\right)^{1 / 2}\right)
    \end{equation}
}

We also use two pairs of metrics, denoted Data-Manifold metrics, based on multi-dimensional representation manifolds \cite{kynkanniemi2019improved,naeem2020reliable}, meaning sets of close neighbors forming a hypersphere.
On graphs, with $B(x,r)$ a ball of center x and radius r, and $\text{NND}_k(X_i)$ the distance between $X_i$, the representation of one graph extracted from the GNN, and the kth nearest neighbor in the embedding space, the manifold is defined as 
\begin{equation}
    \text{manifold}(X_1,\cdots,X_N):= \bigcup_{i=1}^{N} B(X_i,\text{NND}_k(X_i))
\end{equation}
The two pairs of metrics denoted \textit{Precision and Recall}, and \textit{Density and Coverage}, are used to compute \textit{F1 Score}, a metric also accounting for the quality and diversity of the generated sample. \\
\textit{Precision and Recall} \cite{kynkanniemi2019improved} are respectively the proportion of generated samples being in the manifold of the real objects and the proportion of the real objects in the manifold of generated samples.
Finally, \textit{Density and Coverage} \cite{naeem2020reliable} are the proportion of the number of real object manifolds for each generated sample and the proportion of real object manifolds containing at least one generated sample.

\paragraph{Node feature engineering on unattributed Graphs}
Generally, GNNs are very powerful when the graphs are attributed, where the nodes are associated with natural feature vectors corresponding to known attributes. 
With unattributed graphs, we have no information other than the graph's topology given by its adjacency matrix.
In order to compute \textit{FD} using GIN, \cite{ggmiem} used a one-hot vector encoding of node degrees as input node feature vector, as in \cite{xu2019powerful} on social network graphs.
Such feature vectors have been studied and shown to be among the feature initialization that gives overall good performance in graph classification \cite{errica2020fair,cui2021positional}.
Recently, other feature engineering has been explored, using random node features initialization \cite{abboud2021surprising,sato2021random}. 
\cite{sato2021random} notably takes up GIN, adding the ability to generate a random component at each pass through GIN, demonstrating its power on some problems that GIN alone cannot solve. We also implemented this approach alongside the existing node feature initialization approaches explored for generative graph evaluation.

\paragraph{Related Work}
First attempts of importing GAN-related metrics such as the  Fréchet Inception Distance \textit{FID} \cite{heusel2018gans} \textit{Improved Precision and Recall} \cite{naeem2020reliable} were implemented using recent Graph Neural Network techniques such as Graph Isomorphism Network \textit{GIN} \cite{xu2019powerful}, but were only marginally used to implement graph version of existing metrics to work on \cite{ggmiem,Thompson2020BuildingLU}.
Obray and al. \cite{obray2021evaluation} investigated kernel-based evaluation metrics using graph invariants, especially MMD on graph statistics. They show how the kernel choice and fine-tuning of parameters are critical as the metric may become unreliable, giving an unstable ranking of the compared generative model.
Thompson et al. \cite{thompson2022on} recently investigated graph embeddings-based metrics, justifying the potential of using untrained GNN to extract graph embeddings, using similar experiments as in \cite{obray2021evaluation}.
We took \cite{obray2021evaluation} approach for systematically evaluating those graph embeddings-based metrics. Nevertheless, we developed our paper independently from \cite{thompson2022on} as we differentiate on studying different hyperparameter settings and their impact on the behavior of the graph embeddings metrics, such as the choice of node feature initialization and pre-training dataset choice.

\section{GraphRNN \& GRAN}\label{sec:graphrnn-gran}
Both Graph Recurrent Neural Networks \textbf{GraphRNN} \cite{DBLP:journals/corr/abs-1802-08773}, and Graph Recurrent Attention Networks \textbf{GRAN} \cite{DBLP:journals/corr/abs-1910-00760} are autoregressive graph generators. We will then present the autoregressive graph generation problem before detailing how GRAN and GraphRNN work.

\ifthenelse{\boolean{requirement_aaai}}{}{
    \ifthenelse{\boolean{requirement_twocolumn}}{\begin{figure*}[tbh]}{\begin{figure}[tbh]}
        \centering
        \includegraphics[width=\linewidth]{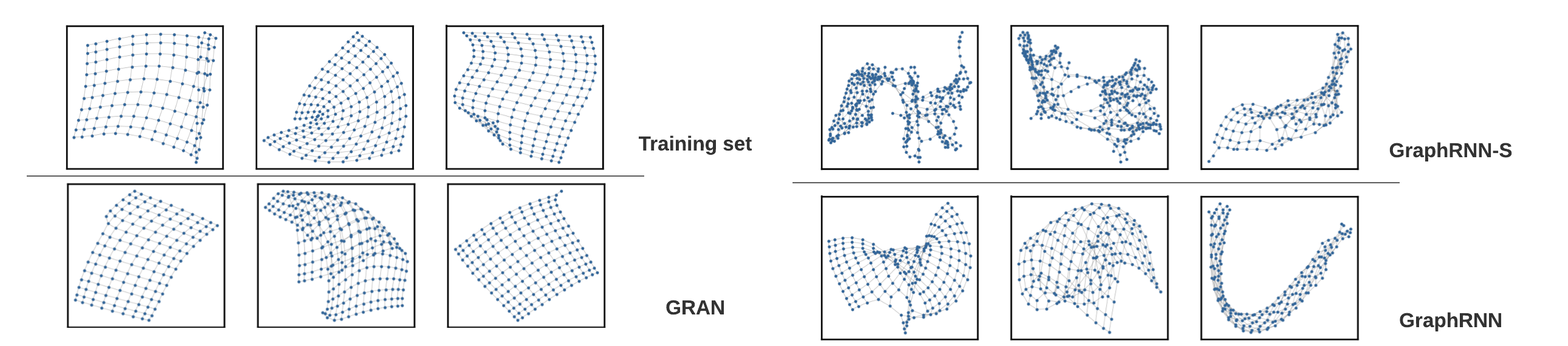}
        \caption{
            Visualisation of generated 2d grid graphs from GRAN, GraphRNN-S and GraphRNN-RNN compared to originally generated grid graphs used as \textit{training set}.
        }
        \label{fig:grid}
        \vspace{-2.5em}
    \ifthenelse{\boolean{requirement_twocolumn}}{\end{figure*}}{\end{figure}}
}

\paragraph{Autoregressive graph generation problem}
We define $G=(V, E)$ as an undirected graph defined by its set of nodes $V$ and edges between nodes $E$.
With the node ordering $\pi$ of $G$, and $V^{\pi}=(v_1,\dots,v_n)$ the ordered set of nodes, we have $A^{\pi}$ the adjacency matrix of G ordered by $\pi$, a $n \times n$ boolean matrix representing G.
Graph generative models learn a probability distribution $P_G$ by learning the sequence $S^{\pi} = (S_{n}^{\pi},..., S_{1}^{\pi})$ with $S_{i}^{\pi}$ corresponding to the i-th row in the adjacency matrix $A^{\pi}$.
We usually solve the problem of graph generation by maximizing the log-likelihood over $log(P_G)$.
Learning to generate rows of adjacency matrices in sequences is difficult because the search space of different adjacency matrices associated to one graph can be huge, up to $n!$, due to the factorial nature of node permutations.
Both \textbf{GRAN} and \textbf{GraphRNN} precisely use chosen node orderings to alleviate this sampling complexity.

Graph Recurrent Neural Networks \textbf{GraphRNN} are based on two RNNs to model the sequence $S^{\pi}$. 
The first RNN is called graph-level RNN, used to save the current state of the generated graph.
The second RNN is called node-level RNN, generating the rows of the adjacency matrix from the current state of the graph given by the graph-level RNN.
Breadth-First-Search (BFS) restricts the maximum node permutations for one graph.
BFS ordering is also used to improve scalability by allowing the model to work on adjacency matrices presenting a known lower graph bandwidth, reducing the model complexity from $n^{2}$ to $n\cdot m$ with $m \le n$.

Graph Recurrent Attention Networks \textbf{GRAN} use attention-based Graph Neural Networks to generate a graph in, at best $O(n)$ autoregressive steps.
Each step generates rows of the adjacency matrix representation of $G$.
GRAN allows for generating multiple rows at once for a quality/time efficiency tradeoff.
Compared to GraphRNN, this model does not save all the previous graph representations and only works on the current graph representation for the autoregressive step.
This model also allows using several sets of node orderings (called canonical node orderings) that could improve generation by having a higher lower bound of the maximum likelihood $\log(P_G)$.
But this comes with a tradeoff between training time and sample quality.

\paragraph{Using several node ordering on GraphRNN and GRAN}
GraphRNN uses BFS to improve its scalability, but this model still suffers from having a relatively high model complexity $O(n^{2})$ and having to deal with very long-term dependencies due to the presence of RNN blocks, which prevents achieving scalability large enough for applications that need it \cite{DBLP:journals/corr/abs-1910-00760}. 
Therefore, the interest in using GraphRNN would be on small graph datasets, where we have already seen an excellent performance from this model \cite{du2021graphgt}.
At this scale of graphs scalability benefits of BFS become irrelevant.
We present the use of Depth-First-Search (DFS) as a node ordering policy, which also makes the problem of graph generation easier by reducing the search space of the graph representation.
However, how much do the performance and quality change when DFS as the node ordering?

GRAN already allows for the possibility of using different node orderings to train the model.
We compare the use of these node orderings on GRAN to accompany the comparative study on GraphRNN.
The main difference between node ordering in both models is that in GRAN the matrix representations of the dataset graphs are fixed at the beginning, so the node ordering function is only used at the beginning of the training.
In GraphRNN, graphs matrix representations change through training because the stochastic node ordering function (e.g., GraphRNN BFS function is computed with a random starting node) is called each time a graph is picked from the dataset during the training loop to give a node ordering that can be different from the one computed previously for the same graph.

\paragraph{Related Graph Generative Models}
Early graph generative models are low parameterized and not powerful enough to model complex dependencies in real-world networks.
Graph generators use deep learning techniques to mimic graphs from real-world data.
Models such as variational auto-encoders \cite{simonovsky2018graphvae,kipf2016variational} or generative adversarial networks \cite{bojchevski2018netgan} have been studied for graphs.
Breakthroughs in quality and scalability happened with autoregressive models like GraphRNN \cite{DBLP:journals/corr/abs-1802-08773}.
Due to their sequential nature, auto-regressive formulations must restrict the space of studied node orderings to make the problem tractable. 
This problem has been studied e.g., in DGMG, and GraphRNN \cite{li2018learning,stier2021deepgg,DBLP:journals/corr/abs-1802-08773}.

\section{Experiments}\label{sec:experiments}
Our study has two concerns:
1/ Studying the behavior of graph embeddings-based metrics using two experiments where we vary several hyperparameters.
2/ See whether GRAN or GraphRNN generates better graphs in terms of quality and how node orderings can impact both models' performances.
All experiments were done using the same Python environment (Python 3.7 + Pytorch 1.8.1).

\ifthenelse{\boolean{requirement_aaai}}{}{
}

\ifthenelse{\boolean{requirement_twocolumn}}{
    \begin{table*}[tbh]
    \centering
    \begin{tabular}{lllll|lll}
        \multicolumn{2}{l}{Dataset $\boxed{B}$ase} & \multicolumn{3}{l}{Pooling $x_{\sum}$} & \multicolumn{3}{l}{Pooling $x_{\mathbin\Vert}$} \\
        & $\boxed\leq^1$ $\downarrow $ & FD  $\downarrow $ & F1 PR $\uparrow $ & F1 DC $\uparrow$ 
        & FD  $\downarrow $ & F1 PR $\uparrow $ & F1 DC $\uparrow$ \\
        \hline

$\ast$   & 1  & -0.0010{\tiny$\pm$0.0007}     & 1.0{\tiny$\pm$0.0}             & 0.994{\tiny$\pm$0.003}
    & \textbf{-0.001{\tiny$\pm$0.001}}       & \textbf{1.0{\tiny$\pm$0.0}}    & \textbf{0.994{\tiny$\pm$0.003}} \\
    \hline

$\vardiamond$         & 2                         & 41004{\tiny$\pm$14957}        & 0.845{\tiny$\pm$0.006}         & 0.79{\tiny$\pm$0.030} 
    &  26639{\tiny$\pm$14495}                 & 0.852{\tiny$\pm$0.010}         & 0.789{\tiny$\pm$0.016} \\

$\clubsuit$-RNN & 3                         & 91512{\tiny$\pm$24215}        & 0.040{\tiny$\pm$0.010}         & 0.019{\tiny$\pm$0.009}
    & 65917{\tiny$\pm$24978}                 & 0.042{\tiny$\pm$0.005}         & 0.026{\tiny$\pm$0.006} \\
$\clubsuit$-S   & 4                         & 319669{\tiny$\pm$94417}       & 0.0{\tiny$\pm$0.0}             & 0.0{\tiny$\pm$0.0}
    & 201977{\tiny$\pm$89750}                & 0.0{\tiny$\pm$0.0}             & 0.0{\tiny$\pm$0.0} \\

    \multicolumn{8}{l}{} \\

        \multicolumn{2}{l}{Dataset $\boxed{L}$adder} & \multicolumn{3}{l}{Pooling $x_{\sum}$} & \multicolumn{3}{l}{Pooling $x_{\mathbin\Vert}$} \\
        & $\boxed\leq^1$ $\downarrow $    & FD  $\downarrow $   & F1 PR   $\uparrow $ & F1 DC   $\uparrow $ \\ \hline

$\ast$   & 1                          & -0.0010{\tiny$\pm$0.0008}      & 1.0{\tiny$\pm$0.0}             & 0.992{\tiny$\pm$0.005}         
    & \textbf{-0.002{\tiny$\pm$0.002}}        & \textbf{1.0{\tiny$\pm$0.0}}    & \textbf{0.990{\tiny$\pm$0.003}} \\
\hline
$\vardiamond$         & 2                          & 30773{\tiny$\pm$23098}         & 0.870{\tiny$\pm$0.041}         & 0.81{\tiny$\pm$0.10} 
    & 10228{\tiny$\pm$11339}                  & 0.879{\tiny$\pm$0.029}         & 0.832{\tiny$\pm$0.069} \\
$\clubsuit$-RNN & 3                          & 127572{\tiny$\pm$29497}        & 0.042{\tiny$\pm$0.015}         & 0.02{\tiny$\pm$0.01}
    & 91113{\tiny$\pm$24230}                  & 0.047{\tiny$\pm$0.005}         & 0.034{\tiny$\pm$0.011} \\
$\clubsuit$-S   & 4                          & 362265{\tiny$\pm$149052}       & 0.001{\tiny$\pm$0.006}         & 0.003{\tiny$\pm$0.001}
    & 212305{\tiny$\pm$78070} & 0.003{\tiny$\pm$0.008}         & 0.0006{\tiny$\pm$0.001} \\

    \multicolumn{8}{l}{} \\

        \multicolumn{2}{l}{Dataset $\boxed{F}$ull} & \multicolumn{3}{l}{Pooling $x_{\sum}$} & \multicolumn{3}{l}{Pooling $x_{\mathbin\Vert}$} \\
        & $\boxed\leq^1$ $\downarrow $ & FD  $\downarrow $       & F1 PR   $\uparrow $ & F1 DC   $\uparrow $  \\  
        \hline

$\ast$   & \textbf{1}              & \textbf{-0.0002{\tiny$\pm$0.0001}} & \textbf{1.0{\tiny$\pm$0.0}}    & \textbf{0.990{\tiny$\pm$0.006}}
    & \textbf{-6e-4{\tiny$\pm$-6e-4}} & \textbf{1.0{\tiny$\pm$0.0}}    & \textbf{0.990{\tiny$\pm$0.007}} \\
    \hline

$\vardiamond$         & 2                       & 737{\tiny$\pm$340}                 & 0.880{\tiny$\pm$0.016}         & 0.847{\tiny$\pm$0.039}
    & 1264{\tiny$\pm$360}             & 0.885{\tiny$\pm$0.0147}        & 0.862{\tiny$\pm$0.029} \\
$\clubsuit$-RNN & 3                       & 45270{\tiny$\pm$16467}             & 0.046{\tiny$\pm$0.006}         & 0.027{\tiny$\pm$0.007} 
    & 59471{\tiny$\pm$9750}           & 0.048{\tiny$\pm$0.001}         & 0.033{\tiny$\pm$0.003} \\
$\clubsuit$-S   & 4                       & 75501{\tiny$\pm$29895}             & 0.0{\tiny$\pm$0.0}             & 0.0{\tiny$\pm$0.0}
    & 91310{\tiny$\pm$19924}          & 0.001{\tiny$\pm$0.006}         & 0.0003{\tiny$\pm$0.001} \\
    \bottomrule
    \multicolumn{5}{l}{\small $\ast$ Re-Sampled, $\boxed\leq^1$ Mean Rank, $\clubsuit$ GraphRNN, $\vardiamond$ GRAN}        
    \end{tabular}

    \caption{
        Metrics for graph embedding pooled with summation $x_{\sum}$ or concatenation $x_{\mathbin\Vert}$ on Gr dataset from 10 independently trained GIN.
        All of the metrics show consistent ranking fidelity with the base ranking hypothesis and stability.
        F1 Score metrics are more stable in regards to the dataset configuration, with low variance compared to FD.
    }
    \label{tab:GINmesSumFull}
    \label{tab:GINmesConcatFull}
\end{table*}

}{
    \ifthenelse{\boolean{requirement_kes}}{
        
    }{
        \begin{table}[tbh]
    \centering
    \begin{tabular}{lllll}
        & \multicolumn{2}{l}{Base dataset configuration} &                     &                     \\ \cline{2-5} 
        & Mean Rank $\downarrow $   & FD  $\downarrow $  & F1 PR   $\uparrow $ & F1 DC   $\uparrow $ \\ \hline
Re-sampled   & 1                         & -0.0010$\pm$0.0007     & 1.0$\pm$0.0             & 0.994$\pm$0.003         \\ \hline
GRAN         & 2                         & 41004$\pm$14957        & 0.845$\pm$0.006         & 0.79$\pm$0.030          \\
GraphRNN-RNN & 3                         & 91512$\pm$24215        & 0.040$\pm$0.010         & 0.019$\pm$0.009         \\
GraphRNN-S   & 4                         & 319669$\pm$94417       & 0.0$\pm$0.0             & 0.0$\pm$0.0            
\end{tabular}

    \begin{tabular}{lllll}
        & \multicolumn{2}{l}{Ladder dataset configuration} &                     &                     \\ \cline{2-5} 
        & Mean Rank $\downarrow $    & FD  $\downarrow $   & F1 PR   $\uparrow $ & F1 DC   $\uparrow $ \\ \hline
Re-sampled   & 1                          & -0.0010$\pm$0.0008      & 1.0$\pm$0.0             & 0.992$\pm$0.005         \\ \hline
GRAN         & 2                          & 30773$\pm$23098         & 0.870$\pm$0.041         & 0.81$\pm$0.10           \\
GraphRNN-RNN & 3                          & 127572$\pm$29497        & 0.042$\pm$0.015         & 0.02$\pm$0.01           \\
GraphRNN-S   & 4                          & 362265$\pm$149052       & 0.001$\pm$0.006         & 0.003$\pm$0.001        
\end{tabular}

    \begin{tabular}{lllll}
        & \multicolumn{2}{l}{Full dataset configuration}    &                     &                      \\ \cline{2-5} 
        & Mean Rank $\downarrow $ & FD  $\downarrow $       & F1 PR   $\uparrow $ & F1 DC   $\uparrow $  \\ \hline
Re-sampled   & \textbf{1}              & \textbf{-0.0002$\pm$0.0001} & \textbf{1.0$\pm$0.0}    & \textbf{0.990$\pm$0.006} \\ \hline
GRAN         & 2                       & 737$\pm$340                 & 0.880$\pm$0.016         & 0.847$\pm$0.039          \\
GraphRNN-RNN & 3                       & 45270$\pm$16467             & 0.046$\pm$0.006         & 0.027$\pm$0.007          \\
GraphRNN-S   & 4                       & 75501$\pm$29895             & 0.0$\pm$0.0             & 0.0$\pm$0.0             
\end{tabular}
    \caption{
        Sum Graph Manifolds metrics on Gr dataset from 10 independently trained GIN.
        All of the metrics show consistent ranking fidelity with the base ranking hypothesis and stability.
        F1 Score metrics are more stable in regards to the dataset configuration, with low variance compared to FD.
    }
    \label{tab:GINmesSumFull}
    \vspace{-3em}
\end{table}

        \begin{table}[tbh]
    \centering
    \begin{tabular}{lllll}
        & \multicolumn{2}{l}{Base dataset configuration} &                     &                      \\ \cline{2-5} 
        & Mean Rank $\downarrow $        & FD  $\downarrow $           & F1 PR   $\uparrow $ & F1 DC   $\uparrow $  \\ \hline
Re-sampled   & \textbf{1}       & \textbf{-0.001$\pm$0.001}       & \textbf{1.0$\pm$0.0}    & \textbf{0.994$\pm$0.003} \\ \hline
GRAN         & 2                & 26639$\pm$14495                 & 0.852$\pm$0.010         & 0.789$\pm$0.016          \\
GraphRNN-RNN & 3                & 65917$\pm$24978                 & 0.042$\pm$0.005         & 0.026$\pm$0.006          \\
GraphRNN-S   & 4                & 201977$\pm$89750                & 0.0$\pm$0.0             & 0.0$\pm$0.0             
\end{tabular}
\begin{tabular}{lllll}
    & \multicolumn{2}{l}{Ladder dataset configuration} &                     &                      \\ \cline{2-5} 
    & Mean Rank $\downarrow $         & FD  $\downarrow $            & F1 PR   $\uparrow $ & F1 DC   $\uparrow $  \\ \hline
Re-sampled   & \textbf{1}        & \textbf{-0.002$\pm$0.002}        & \textbf{1.0$\pm$0.0}    & \textbf{0.990$\pm$0.003} \\ \hline
GRAN         & 2                 & 10228$\pm$11339                  & 0.879$\pm$0.029         & 0.832$\pm$0.069          \\
GraphRNN-RNN & 3                 & 91113$\pm$24230                  & 0.047$\pm$0.005         & 0.034$\pm$0.011          \\
GraphRNN-S   & 4                 & 212305$\pm$78070                 & 0.003$\pm$0.008         & 0.0006$\pm$0.001        
\end{tabular}
\begin{tabular}{lllll}
    & \multicolumn{2}{l}{Full dataset configuration} &                     &                      \\ \cline{2-5} 
    & Mean Rank $\downarrow $ & FD  $\downarrow $    & F1 PR   $\uparrow $ & F1 DC   $\uparrow $  \\ \hline
Re-sampled   & \textbf{1}              & \textbf{-6e-4$\pm$-6e-4} & \textbf{1.0$\pm$0.0}    & \textbf{0.990$\pm$0.007} \\ \hline
GRAN         & 2                       & 1264$\pm$360             & 0.885$\pm$0.0147        & 0.862$\pm$0.029          \\
GraphRNN-RNN & 3                       & 59471$\pm$9750           & 0.048$\pm$0.001         & 0.033$\pm$0.003          \\
GraphRNN-S   & 4                       & 91310$\pm$19924          & 0.001$\pm$0.006         & 0.0003$\pm$0.001        
\end{tabular}
    \caption{
        Concatenation Graph Manifolds metrics on Gr dataset from 10 independently trained GIN.
        All of the metrics show consistent ranking fidelity with the base ranking hypothesis and stability.
        F1 Score metrics are more stable in regards to the dataset configuration, with low variance compared to FD.
     }
    \label{tab:GINmesConcatFull}
    \vspace{-2em}
\end{table}

    }
}
\ifthenelse{\boolean{requirement_twocolumn}}{\begin{table}[tbh]
    \centering
    \begin{tabular}{lllll}
        \multicolumn{5}{l}{With \textbf{constant} feature} \\
        & ${\boxed\leq^1}$  & FD  $\downarrow$ & F1 PR $\uparrow $ & F1 DC $\uparrow$ \\
        \hline
        $\ast$ & \textbf{1} & \textbf{-8e-5{\tiny$\pm$5e-5}} & \textbf{1.0{\tiny$\pm$0.0}} & \textbf{0.975{\tiny$\pm$0.010}} \\
        \hline
        $\vardiamond$ & 2  & 211{\tiny$\pm$29}  & 0.973{\tiny$\pm$0.015} & 0.973{\tiny$\pm$0.015} \\
        $\clubsuit$-R & 4 & 30904{\tiny$\pm$2275} & 0.007{\tiny$\pm$0.010} & 0.052{\tiny$\pm$0.0001} \\
        $\clubsuit$-S & 3 & 26335{\tiny$\pm$2016} & 0.60{\tiny$\pm$0.027} & 0.056{\tiny$\pm$0.010} \\
        \\
        \multicolumn{5}{l}{With \textbf{random} feature (rGIN)} \\
        $\ast$ & \textbf{1} & \textbf{0.87{\tiny$\pm$0.82}} & \textbf{0.988{\tiny$\pm$0.006}} & \textbf{1.013{\tiny$\pm$0.021}} \\
        \hline
        $\vardiamond$ & 2 & 61{\tiny$\pm$13} & 0.967{\tiny$\pm$0.013} & 0.989{\tiny$\pm$0.036} \\
        $\clubsuit$-R & 4 & 8989{\tiny$\pm$1076} & 0.017{\tiny$\pm$0.006} & 0.058{\tiny$\pm$0.005} \\
        $\clubsuit$-S & 3 & 8062{\tiny$\pm$925} & 0.535{\tiny$\pm$0.074} & 0.065{\tiny$\pm$0.021} \\
        \bottomrule
        \multicolumn{5}{l}{$\ast$ Re-Sampled, \small ${\boxed\leq^1}$ Mean Rank}
    \end{tabular}
    \caption{
        Graph embedding metrics (extracted by concatenation of node embedding withing GIN) of Gr dataset on 10 independentely trained GIN ($\boxed{F}$ull dataset), using rGIN and constant features.
        While the metrics were stable through GIN training repetition, their ranking fidelity in regards to the base ranking hypothesis is not perfect, ranking the GraphRNN-S better than the GraphRNN-RNN generated grid graphs.
    }
    \label{tab:GINmesranda}
\end{table}
}{\begin{table}[tbh]
    \centering
    \begin{tabular}{lllll}
        \multicolumn{5}{l}{With constant feature} \\
        \hline
        & Mean Rank  & FD  $\downarrow $   & F1 PR   $\uparrow $ & F1 DC   $\uparrow $  \\ \hline
    Re-sampled   & \textbf{1} & \textbf{-8e-5$\pm$5e-5} & \textbf{1.0$\pm$0.0}    & \textbf{0.975$\pm$0.010} \\ \hline
    GRAN         & 2          & 211$\pm$29              & 0.973$\pm$0.015         & 0.973$\pm$0.015          \\
    GraphRNN-RNN & 4          & 30904$\pm$2275          & 0.007$\pm$0.010         & 0.052$\pm$0.0001         \\
    GraphRNN-S   & 3          & 26335$\pm$2016          & 0.60$\pm$0.027          & 0.056$\pm$0.010         
    \end{tabular}
    \begin{tabular}{lllll}
        \multicolumn{5}{l}{With random feature (rGIN)} \\
        \hline
        & Mean Rank          & FD  $\downarrow $         & F1 PR   $\uparrow $  & F1 DC   $\uparrow $  \\ \hline
    Re-sampled   & \textbf{1}         & \textbf{0.87$\pm$0.82}        & \textbf{0.988$\pm$0.006} & \textbf{1.013$\pm$0.021} \\ \hline
    GRAN         & 2                  & 61$\pm$13                     & 0.967$\pm$0.013          & 0.989$\pm$0.036          \\
    GraphRNN-RNN & 4                  & 8989$\pm$1076                 & 0.017$\pm$0.006          & 0.058$\pm$0.005          \\
    GraphRNN-S   & 3                  & 8062$\pm$925                  & 0.535$\pm$0.074          & 0.065$\pm$0.021         
    \end{tabular}
    \caption{
        Graph manifolds metrics of Gr dataset on 10 independentely trained GIN ($\boxed{F}$ull dataset), using rGIN and constant features.
        While the metrics were able to keep consistent results through GIN training repetition, they kept unconsistent ranking results in regards to the base hypothesis, ranking the GraphRNN-S better than the GraphRNN-RNN generated grid graphs.
    }
    \label{tab:GINmesranda}
\end{table}
}

\paragraph{Datasets}
For both parts, we focus on synthetic datasets for two reasons:
First, some synthetic graphs datasets are based on mathematical models such as Barabasi-Albert \textbf{BA} and Watts-Strogatz \textbf{WS} models or have directly identifiable topologies such as community datasets.
Finally, using such synthetic datasets enable us to directly compute benchmarks for a \textit{perfect} generation being the synthetic graph generator itself, which is further explained in the Results section.

The following datasets are used for our experiment: (1) \textbf{BA}: 500 Barabasi-Albert graphs \cite{Barabasi509} with $100 \le n \le 200 $  using the B-A model, using the integrated networkx.barabasi\_albert\_graph function \cite{hagberg2008exploring}, with parameter $k=4$.

\noindent (2) \textbf{WS}: 500 Watts-Strogatz graphs \cite{watts1998collective} with $100 \le n \le 200$  using the W-S model, where each node is linked to its four nearest neighbor and the probability of rewiring is 0.1.

\noindent (3) \textbf{C2L}: 500 2-community graphs using the synthetic community generator from \cite{DBLP:journals/corr/abs-1802-08773}, where communities are formed with Erdos-Renyi graphs of $6 \le  n_{community} \le 10 $  and $p=0.7$ from which $0.1 \times n$ edges between communities are added uniformly at random.

\noindent (4) \textbf{C2S}: 500 2-community graphs using the synthetic community generator from \cite{DBLP:journals/corr/abs-1802-08773}, where communities are formed with Erdos-Renyi graphs of $30 \le n_{community} \le 80 $  and $p=0.3$ from which $0.05 \times n$ edges between communities are added uniformly at random.

\noindent (5) \textbf{Gr}: 100 2D grid graphs with $100 \le n \le 400 $ using the integrated networkx.2d\_grid\_graph function.
 
\noindent (6) \textbf{Ld}: 500 ladder graphs with $100 \le n \le 200 $ using the integrated networkx.ladder\_graph function.

\noindent (7) \textbf{ER}: we create this dataset with one of the previously mentioned graph datasets, where an ER graph is created with $n = max(n_{g},\text{label}(g) = i)$ and parameter $p = \frac{e_{g}}{n_{g}^{2}}$, this dataset contains the same amount of graph as the linked graph dataset (e.g., we would generate 500 ER graphs "equivalent" to the \textbf{BA} dataset, or 100 ER graphs "equivalent" to the \textbf{Gr} dataset)

\subsection{Testing Evaluation Metrics with Graph Embeddings}
We evaluate the behavior of graph embeddings-based metrics computed using Graph Neural Networks trained on a graph classification task.
We want to observe how changes in the training process, such as the initialization of node feature vectors or the choice of the training dataset, would impact the behavior of the resulting metrics and gain insights into optimizing the training process to get more expressive metrics.

\paragraph{Experiment Description}
In their work, Obray and al. discussed criteria for evaluating comparison metrics, among which we find expressiveness, where the ideal comparison metric can rank graph samples effectively \cite{obray2021evaluation}.   
To evaluate expressiveness, we use two experiments:
First, in the \textbf{Perturbation experiment}, we compare a set of graphs to a copy set but perturbed with interpolation between those graphs and rewired ER-graphs as in \cite{DBLP:journals/corr/abs-1802-08773,obray2021evaluation}.
We expect the metrics to react monotonically to those perturbations, keeping the similarity rank of each perturbed graph set with the reference set.
We also look for metric stability if those metrics keep the low variance regarding experiment repetition, which includes pre-training and metrics computation.
Second, in the \textbf{Grid experiment}, we compare Gr graphs from the real data distribution with ones sampled from GRAN and GraphRNN, which gives a closer outlook on using those metrics in an actual research workflow.
We expect the metrics based on graph embeddings to rank these graph sets consistently with our visual inspection.

\paragraph{Experiment Parameters}
In both experiments, we varied training training hyperparameters of the graph classifier: 
\textit{1st/} Node feature vector initialization for pre-training and evaluation. 
We compare three feature initializations:
One-hot degree encoding, Constant node encoding, and GIN-random \cite{sato2021random}, where each node feature is generated by sampling a random integer each time the model is called during training and inference time.
\textit{2nd/} The graph dataset used for pre-training.
We concatenated the graph data to work with three graph classification datasets:
(1) The \textbf{base dataset} $\boxed{B}$ consists of the graph datasets BA, WS, C2L, C2S and Gr.
(2) The \textbf{ladder dataset} $\boxed{L}$ takes the base dataset and add Ld graphs set.
(3) The \textbf{full dataset} $\boxed{F}$ use the ladder dataset and add ER graph sets for each already existing graph set.
\textit{3rd/} The GIN graph embedding function, whether summation or concatenation.

\ifthenelse{\boolean{requirement_twocolumn}}{
    \begin{table}[]
    \centering
\resizebox{\columnwidth}{!}{
    \begin{tabular}{|l|l|l|l|l|}
    \hline
    & Training dataset setup & $\rho$@FD            & $\rho$@F1-PR & $\rho$@F1-DC \\
    \hline
    \multirow{2}{*}{Grid graphs} & Dataset $\boxed{B}$ase         & 0.72{\tiny$\pm$0.021}   & -0.5{\tiny$\pm$0.0}   & -0.5{\tiny$\pm$0.0} \\
    \cline{2-5} 
    & Dataset $\boxed{F}$ull           & 1.0{\tiny$\pm$0.0}   & -0.5{\tiny$\pm$0.0}   & -0.5{\tiny$\pm$0.0} \\
    \hline
    \multirow{2}{*}{BA graphs}  & Dataset $\boxed{B}$ase          & 0.99{\tiny$\pm$1e-5}  & -0.90{\tiny$\pm$0.001} & -0.92{\tiny$\pm$0.002} \\
    \cline{2-5} 
    & Dataset $\boxed{F}$ull           & 1.00{\tiny$\pm$0.0} & -0.79{\tiny$\pm$0.002} & -0.81{\tiny$\pm$0.006} \\
    \hline
    \multirow{2}{*}{WS graphs}  & Dataset $\boxed{B}$ase          & 0.87{\tiny$\pm$0.077} & -0.67{\tiny$\pm$0.0} & -0.67{\tiny$\pm$0.0} \\
    \cline{2-5} 
   & Dataset $\boxed{F}$ull          & 0.92{\tiny$\pm$0.015}  & -0.68{\tiny$\pm$0.001}  & -0.68{\tiny$\pm$0.001}  \\
   \hline
    \end{tabular}
}
\caption{
            Ablation study on training dataset setup and Evaluation metrics behavior. Mean Spearman rank correlation and variance, collected for FD, F1-Score PR and F1-Score DC on 10 runs. The ranking correlation coefficient were computed by measuring 10 values corresponding to increasing perturbation of the BA, WS and Grid datasets.
        }
        \label{tab:rank-perturb}
    \end{table}

    \begin{table}[]
    \centering
\resizebox{\columnwidth}{!}{
        \begin{tabular}{|l|l|l|l|}
        \hline
        Node feature initialization & $\rho$@FD          & $\rho$@F1-PR        & $\rho$@F1-DC        \\ \hline
        Constant                    & 1.0{\tiny$\pm$0.0} & -0.5{\tiny$\pm$0.0} & -0.5{\tiny$\pm$0.0} \\ \hline
        Random                      & 1.0{\tiny$\pm$0.0} & -0.5{\tiny$\pm$0.0} & -0.5{\tiny$\pm$0.0} \\ \hline
        Degree one-hot encoding     & 1.0{\tiny$\pm$0.0} & -0.5{\tiny$\pm$0.0} & -0.5{\tiny$\pm$0.0} \\ \hline
        \end{tabular}
}
\caption{
    Ablation study on node feature initialization and Evaluation metrics behavior. Mean Spearman rank correlation and variance, collected for FD, F1-Score PR and F1-Score DC on 10 runs. The ranking correlation coefficient were computed by measuring 10 values corresponding to increasing perturbation of the Grid dataset.
        }
        \label{tab:rank-perturb-feature}
    \end{table}

    \begin{table*}[tbh]
    \centering
\resizebox{\textwidth}{!}{
    \begin{tabular}{ll||lll||lll||lll||lll}
    & & \multicolumn{3}{|l||}{WS} & \multicolumn{3}{|l||}{BA} & \multicolumn{3}{|l||}{C2L} & \multicolumn{3}{l}{C2S} \\
    \cline{3-14} 
    &  & Deg & Clust & 4-orbits & Deg & Clust & 4-orbits & Deg & Clust & 4-orbits & Deg & Clust & 4-orbits \\
    \hline
    $\ast$ &  & \textbf{2.718e-6} & \textbf{0.0013} & \textbf{9.780e-6} & \textbf{3.822e-5} & \textbf{0.0035} & \textbf{0.0047} & \textbf{0.0004} & \textbf{0.0036} & \textbf{0.0039} & \textbf{0.0005} & \textbf{0.0037} & \textbf{0.0006} \\
    \hline
    E-R  &  & 0.3412 & 1.3495 & 0.1904  & 0.1074  & 0.6967   & 0.6889 & 0.1481 & 0.2562  & 0.1042 & 0.1715  & 0.1603  & 0.2597 \\
    \hline
    $\vardiamond$  & BFS  & 0.0038   & 0.4425  & 0.0468  & \textbf{0.0199} & \textbf{0.0534} & \textbf{0.0360} &  0.0052 & 0.0189  & 0.0070 & \textbf{0.0009} & 0.0441  & \textbf{0.0074} \\
     & DFS & 0.0054  & 0.1818  & 0.0043 & 0.0339 & 0.0751 & 0.0436 & \textbf{0.0012} & 0.0190  & 0.0057  & 0.0022 & \textbf{0.0374} & 0.0087 \\
    & DegDesc & 0.0180  & 0.5491  & 0.0511 & 0.0706 & 0.1045  & 0.0553 & 0.0262  & 0.0236  & 0.0073 & 0.0038  & 0.0417  & 0.0138 \\
    & Kcore  & 0.0208  & 0.4999   & 0.0232 &  0.0587  & 0.0992  & 0.0379  & 0.0082  & 0.0215  & 0.0070 & 0.0038 & 0.0483  & 0.0405 \\
    & Default & \textbf{0.0001}  & \textbf{0.0752} & \textbf{0.0046}  & 0.1017 & 0.0967  & 0.0367 &  0.0042 & \textbf{0.0115} & \textbf{0.0054} & 0.0080  & 0.0473  & 0.0801 \\
    \hline
    $\clubsuit$ & BFS   & 0.0986  & \textbf{0.6891} & 0.1641  & 0.1083  & \textbf{0.3152} & 0.1525 & 0.0136 & 0.0400  & \textbf{0.0068} & 0.0085  & 0.0486  & 0.0088 \\
    & BFS-Max & 0.0873   & 0.7388  & 0.2242 & 0.1280  & 0.3378 & \textbf{0.1377}  & \textbf{0.0075} & \textbf{0.0346} & 0.0075 & 0.0041  & \textbf{0.0472} & 0.0103 \\
    & DFS  & \textbf{0.0780}   & 0.7094  & \textbf{0.1508} & \textbf{0.0342} & 0.3234 & 0.1690  & 0.0171  & 0.0524  & 0.0083 & \textbf{0.0027} & \textbf{0.0472} & \textbf{0.0051} \\
    & Uniform & 0.1801  & 1.1016  & 1.7263   & 0.0937   & 0.3775  & 0.1531  & 0.1594  & 0.2463  & 0.0103 & 0.0431  & 0.0503 & 1.193 \\
    \bottomrule
    \end{tabular}
}

    \caption{
        Ablation study on the node ordering of GRAN $\vardiamond$ and GraphRNN $\clubsuit$ on BA-, WS-, C2L- and C2S-graphs with MMD values.
        Tiny metric values yields better generation quality.
        Overall GRAN gets lower MMD metric values on both WS and BA datasets, while the choice of node ordering clearly influence results.
        On GraphRNN, DFS ordering give comparable results to the vanilla BFS orderings.
        While on C3L graphs GRAN gets lower MMD values, GraphRNN gets similar results except for the uniform node ordering.
        On C2S graphs both model also gave similar results, with a slight advantage for GRAN.
        GraphRNN-DFS gave comparable results to the other GraphRNN node ordering variants.
    }
    \label{tab:wsba}
    \label{tab:com}
\end{table*}

\begin{table*}[h!]
    \centering
\resizebox{\textwidth}{!}{
\begin{tabular}{ll||l|l|l||l|l|l||l|l|l||l|l|l}
    \multicolumn{2}{l||}{} & \multicolumn{3}{l||}{WS} & \multicolumn{3}{l||}{BA} & \multicolumn{3}{l||}{C2L} & \multicolumn{3}{l}{C2S} \\
    \cline{3-14}
    & & FD $\downarrow$ & F1 PR $\uparrow$ & F1 DC $\uparrow$ & FD $\downarrow$ & F1 PR $\uparrow$ & F1 DC $\uparrow$ & FD $\downarrow$ & F1 PR $\uparrow$ & F1 DC $\uparrow$ & FD $\downarrow$ & F1 PR $\uparrow$ & F1 DC $\uparrow$\\
    \hline
* & & 4.450 & 0.9799 & 1.001 & 13.89 & 0.9540 & 0.9567 & 1487 & 0.9799 & 1.011 & 6.80 & 0.424 & 0.375 \\
    \hline
$\vardiamond$ & BFS & 1332 & 0.4297 & 0.1344 & \textbf{2399} & \textbf{0.2132} & \textbf{0.0771} & 47782 & 0.8881 & 0.6984 & 16.29 & \textbf{0.8207} & 0.3466\\
    & DFS & 1708 & 0.4546 & 0.2341 & 3323 & 0.1431 & 0.0315 & 21005 & 0.9201 & 0.7523 & 26.59 & 0.7951 & \textbf{0.3557}\\
    & DegDes & 2189 & 0.2535 & 0.0646 & 17275 & 0.0366 & 0.0065 & 54148 & 0.8300 & 0.5670 & 31.41 & 0.7335 & 0.2813\\
    & Kcore & 1843 & 0.2535 & 0.0771 & 7057 & 0.0514 & 0.0101 & 22293 & 0.7822 & 0.5559 & 25.18 & 0.7415 & 0.2764\\
    & Default & \textbf{308} & \textbf{0.6964} & \textbf{0.4212} & 14364 & 0.0068 & 0.0011 & 45217 & 0.8819 & 0.7630 & 74.07 & 0.6491 & 0.2679\\
    \hline
$\clubsuit$ & BFS & 44503 & 0.0 & 0.0 & 20077 & 0.0 & 0.0 & 12751 & 0.9588 & 0.7883 & 44.69 & 0.2208 & 0.0549\\
    & BFS-Max & 37641 & 0.0 & 0.0 & 30684 & 0.0 & 0.0 & \textbf{5312} & \textbf{0.9678} & \textbf{0.8590} & 61.37 & 0.2305 & 0.0552\\
    & DFS & 22886 & 0.0 & 0.0 & 9762 & 0.0 & 0.0 & 12946 & 0.9242 & 0.7411 & \textbf{13.94} & 0.8060 & 0.3336\\
    & Uniform & 72402 & 0.0 & 0.0 & 28248 & 0.0 & 0.0 & 794711 & 0.1463 & 0.0976 & 182.1 & 0.5762 & 0.1847\\
    \hline
\end{tabular}
}
    \caption{
        Graph embedding metrics for re-sampled graphs of a graph type are denoted in the first row with *.
        Reported pretrained Graph manifolds-based metrics on GRAN ($\vardiamond$) and GraphRNN ($\clubsuit$) for BA and WS dataset, using one-hot degree features initialization and full training dataset configuration, with graph embeddings pooled with concatenation $x_{\mathbin\Vert}$.
        Lower FD values and higher F1 Score (PR and DC) yields better generation quality. We observe a clear difference in generation quality, with GRAN generating graph with higher quality compared to the one generated by GraphRNN on both datasets.
        Those graph manifolds metrics gave similar results to the MMD-based metrics.
    }
    \label{tab:GINmes}
\end{table*}

}{
        \begin{table}[tbh]
    \centering
    \resizebox{\textwidth}{!}{%
    \begin{tabular}{llllllllll}
    \cline{3-10}
            &                & \multicolumn{4}{l}{WS}                                         & \multicolumn{4}{l}{BA}                                     \\ \cline{3-10} 
            &                & Deg               & Clust           & 4-orbits          & Spec            & Deg               & Clust           & 4-orbits        & Spec            \\ \hline
    Re-sampled &                & \textbf{2.718e-6} & \textbf{0.0013} & \textbf{9.780e-6} & \textbf{0.0003} & \textbf{3.822e-5} & \textbf{0.0035} & \textbf{0.0047} & \textbf{0.0001} \\ \hline
    E-R        &                & 0.3412            & 1.3495          & 0.1904            & 0.1436          & 0.1074            & 0.6967          & 0.6889          & 0.0564          \\ \hline
    GRAN       & BFS            & 0.0038            & 0.4425          & 0.0468            & 0.0282          & \textbf{0.0199}   & \textbf{0.0534} & \textbf{0.0360} & 0.0023          \\
            & DFS            & 0.0054            & 0.1818          & 0.0043            & 0.0035          & 0.0339            & 0.0751          & 0.0436          & \textbf{0.0012} \\
            & Degree Descent & 0.0180                             & 0.5491                           & 0.0511                             & 0.0090           & 0.0706            & 0.1045          & 0.0553          & 0.0015          \\
            & Kcore          & 0.0208            & 0.4999          & 0.0232            & 0.0091          & 0.0587            & 0.0992          & 0.0379          & 0.0018          \\
            & Default        & \textbf{0.0001}   & \textbf{0.0752} & \textbf{0.0046}   & \textbf{0.0023} & 0.1017            & 0.0967          & 0.0367          & 0.0031          \\ \hline
    GraphRNN   & BFS            & 0.0986            & \textbf{0.6891} & 0.1641            & 0.0250          & 0.1083            & \textbf{0.3152} & 0.1525          & 0.0184          \\
            & BFS-Max        & 0.0873            & 0.7388          & 0.2242            & 0.0277          & 0.1280            & 0.3378          & \textbf{0.1377} & 0.0150          \\
            & DFS            & \textbf{0.0780}   & 0.7094          & \textbf{0.1508}   & \textbf{0.0203} & \textbf{0.0342}   & 0.3234          & 0.1690          & \textbf{0.0090} \\
            & Uniform        & 0.1801            & 1.1016          & 1.7263            & 0.0711          & 0.0937            & 0.3775          & 0.1531          & 0.0105          \\ \hline
    \end{tabular}%
    }
    \caption{
        Ablation study on Node Ordering of GRAN $\vardiamond$ and GraphRNN $\clubsuit$ on BA and WS graphs, using reported MMD values. Lower values yields better generation quality. Overall GRAN gets lower MMD metrics vazlues on both WS and BA datasets, while the choice of node ordering clearly influence results. On GraphRNN, DFS ordering give comparable results to the vanilla BFS orderings.
    }
    \label{tab:wsba}
\end{table}

        \begin{table}[tbh]
\centering
\resizebox{\textwidth}{!}{%
\begin{tabular}{llllllllll}
\cline{3-10}
           &                & \multicolumn{4}{l}{C2L}                                        & \multicolumn{4}{l}{C2S}                                \\ \cline{3-10} 
           &                & Deg             & Clust           & 4-orbits        & Spec            & Deg             & Clust           & 4-orbits        & Spec            \\ \hline
Re-sampled &                & \textbf{0.0004} & \textbf{0.0036} & \textbf{0.0039} & \textbf{0.0001} & \textbf{0.0005} & \textbf{0.0037} & \textbf{0.0006} & \textbf{0.0011} \\ \hline
E-R        &                & 0.1481          & 0.2562          & 0.1042          & 0.0452          & 0.1715          & 0.1603          & 0.2597          & 0.0730          \\ \hline
GRAN       & BFS            & 0.0052          & 0.0189          & 0.0070          & 0.0017          & \textbf{0.0009} & 0.0441          & \textbf{0.0074} & 0.0223          \\
           & DFS            & \textbf{0.0012} & 0.0190          & 0.0057          & \textbf{0.0007} & 0.0022          & \textbf{0.0374} & 0.0087          & \textbf{0.0108} \\
           & Degree Descent & 0.0262          & 0.0236          & 0.0073          & 0.0029          & 0.0038          & 0.0417          & 0.0138          & 0.0215          \\
           & Kcore          & 0.0082          & 0.0215          & 0.0070          & 0.0011          & 0.0038          & 0.0483          & 0.0405          & 0.0263          \\
           & Default        & 0.0042          & \textbf{0.0115} & \textbf{0.0054} & 0.0010          & 0.0080          & 0.0473          & 0.0801          & 0.0248          \\ \hline
GraphRNN   & BFS            & 0.0136          & 0.0400          & \textbf{0.0068} & 0.0092          & 0.0085          & 0.0486          & 0.0088          & 0.0411          \\
           & BFS-Max        & \textbf{0.0075} & \textbf{0.0346} & 0.0075          & \textbf{0.0050} & 0.0041          & \textbf{0.0472} & 0.0103          & 0.0400          \\
           & DFS            & 0.0171          & 0.0524          & 0.0083          & 0.0068          & \textbf{0.0027} & \textbf{0.0472} & \textbf{0.0051} & \textbf{0.0196} \\
           & Uniform        & 0.1594          & 0.2463          & 0.0103          & 0.0079          & 0.0431          & 0.0503          & 1.193           & 0.0478          \\ \hline
\end{tabular}%
}
\caption{
    Ablation study on Node Ordering of GRAN and GraphRNN on C2L and C2S graphs, using reported MMD values.
    Lower values yields better generation quality.
    While on C3L graphs GRAN gets lower MMD values, GraphRNN gets similar results except for the Uniform ordering.
    On C2S graphs both model also gave similar results, with a slight advantage for GRAN.
    GraphRNN-DFS gave comparable results to the other GraphRNN node ordering variants.
}
\label{tab:com}
\end{table}

    
\begin{table}[tbh]
    \centering
    \resizebox{\textwidth}{!}{

\begin{tabular}{ll||l|l|l||l|l|l||l|l|l||l|l|l}
    \multicolumn{2}{l}{} & \multicolumn{3}{l||}{BA} & \multicolumn{3}{l||}{WS} & \multicolumn{3}{l||}{C2S} & \multicolumn{3}{l}{C2L} \\
    \cline{3-14}
    & & FD $\downarrow$ & F1 PR $\uparrow$ & F1 DC $\uparrow$ & FD $\downarrow$ & F1 PR $\uparrow$ & F1 DC $\uparrow$ & FD $\downarrow$ & F1 PR $\uparrow$ & F1 DC $\uparrow$ & FD $\downarrow$ & F1 PR $\uparrow$ & F1 DC $\uparrow$\\
    \hline
$\varheart$ & & 13.89 & 0.9540 & 0.9567 & 4.450 & 0.9799 & 1.001 & 6.80 & 0.424 & 0.375 & 1487 & 0.9799 & 1.011 \\
    \hline
$\vardiamond$ & BFS & \textbf{2399} & \textbf{0.2132} & \textbf{0.0771} & 1332 & 0.4297 & 0.1344 & 16.29 & \textbf{0.8207} & 0.3466 & 47782 & 0.8881 & 0.6984\\
    & DFS & 3323 & 0.1431 & 0.0315 & 1708 & 0.4546 & 0.2341 & 26.59 & 0.7951 & \textbf{0.3557} & 21005 & 0.9201 & 0.7523\\
    & DegDes & 17275 & 0.0366 & 0.0065 & 2189 & 0.2535 & 0.0646 & 31.41 & 0.7335 & 0.2813 & 54148 & 0.8300 & 0.5670\\
    & Kcore & 7057 & 0.0514 & 0.0101 & 1843 & 0.2535 & 0.0771 & 25.18 & 0.7415 & 0.2764 & 22293 & 0.7822 & 0.5559\\
    & Default & 14364 & 0.0068 & 0.0011 & \textbf{308} & \textbf{0.6964} & \textbf{0.4212} & 74.07 & 0.6491 & 0.2679 & 45217 & 0.8819 & 0.7630\\
    \hline
$\clubsuit$ & BFS & 20077 & 0.0 & 0.0 & 44503 & 0.0 & 0.0 & 44.69 & 0.2208 & 0.0549 & 12751 & 0.9588 & 0.7883\\
    & BFS-Max & 30684 & 0.0 & 0.0 & 37641 & 0.0 & 0.0 & 61.37 & 0.2305 & 0.0552 & \textbf{5312} & \textbf{0.9678} & \textbf{0.8590}\\
    & DFS & 9762 & 0.0 & 0.0 & 22886 & 0.0 & 0.0 & \textbf{13.94} & 0.8060 & 0.3336 & 12946 & 0.9242 & 0.7411\\
    & Uniform & 28248 & 0.0 & 0.0 & 72402 & 0.0 & 0.0 & 182.1 & 0.5762 & 0.1847 & 794711 & 0.1463 & 0.0976\\
    \hline
\end{tabular}

    }
    \caption{
        Graph embedding metrics for re-sampled graphs of a graph type are denoted in the first row with $\varheart$.
        Reported pretrained Graph manifolds-based metrics on GRAN ($\vardiamond$) and GraphRNN ($\clubsuit$) for BA and WS dataset.
        Those graph manifolds metrics give similar observation with the evaluation using MMD-based metrics.
    }
    \label{tab:GINmes}
\end{table}

    \ifthenelse{\boolean{article_extended}}{
        
\begin{table}[tbh]
    \centering
    \begin{minipage}{.55\linewidth}
    \resizebox{\linewidth}{!}{\begin{tabular}{|l|l|l|l|l|}
        \hline
        & Train Data & $\rho$@FD            & $\rho$@F1-PR & $\rho$@F1-DC \\
        \hline
        \multirow{2}{*}{Grid graphs} & $\boxed{B}$ase         & 0.72{\tiny$\pm$0.021}   & -0.5{\tiny$\pm$0.0}   & -0.5{\tiny$\pm$0.0} \\
        \cline{2-5} 
        & $\boxed{F}$ull           & 1.0{\tiny$\pm$0.0}   & -0.5{\tiny$\pm$0.0}   & -0.5{\tiny$\pm$0.0} \\
        \hline
        \multirow{2}{*}{BA graphs}  & $\boxed{B}$ase          & 0.99{\tiny$\pm$1e-5}  & -0.90{\tiny$\pm$0.001} & -0.92{\tiny$\pm$0.002} \\
        \cline{2-5} 
        & $\boxed{F}$ull           & 1.00{\tiny$\pm$0.0} & -0.79{\tiny$\pm$0.002} & -0.81{\tiny$\pm$0.006} \\
        \hline
        \multirow{2}{*}{WS graphs}  & $\boxed{B}$ase          & 0.87{\tiny$\pm$0.077} & -0.67{\tiny$\pm$0.0} & -0.67{\tiny$\pm$0.0} \\
    \cline{2-5} 
        & $\boxed{F}$ull          & 0.92{\tiny$\pm$0.015}  & -0.68{\tiny$\pm$0.001}  & -0.68{\tiny$\pm$0.001}  \\
        \hline
    \end{tabular}}
    \caption{
            Ablation study on training dataset setup and Evaluation metrics behavior. Mean Spearman rank correlation and variance, collected for FD, F1-Score PR and F1-Score DC on 10 runs. The ranking correlation coefficient were computed by measuring 10 values corresponding to increasing perturbation of the BA, WS and Grid datasets.
    }
    \label{tab:rank-perturb}
    \end{minipage}
    \quad
    \begin{minipage}{.41\linewidth}
    \resizebox{\linewidth}{!}{\begin{tabular}{|l|l|l|l|}
        \hline
        Init & $\rho$@FD          & $\rho$@F1-PR        & $\rho$@F1-DC        \\ \hline
        Constant                    & 1.0{\tiny$\pm$0.0} & -0.5{\tiny$\pm$0.0} & -0.5{\tiny$\pm$0.0} \\ \hline
        Random                      & 1.0{\tiny$\pm$0.0} & -0.5{\tiny$\pm$0.0} & -0.5{\tiny$\pm$0.0} \\ \hline
        DOE     & 1.0{\tiny$\pm$0.0} & -0.5{\tiny$\pm$0.0} & -0.5{\tiny$\pm$0.0} \\ \hline
    \end{tabular}}
    \caption{
        Ablation study on node feature initialization and Evaluation metrics behavior. Mean Spearman rank correlation and variance, collected for FD, F1-Score PR and F1-Score DC on 10 runs. The ranking correlation coefficient were computed by measuring 10 values corresponding to increasing perturbation of the Grid dataset.
    }
    \label{tab:rank-perturb-feature}
    \end{minipage}
\end{table}

    }
}

\paragraph{Experimental Setup}
We follow \cite{ggmiem} by using GIN \cite{xu2019powerful} as our primary graph classifier. 
The official implementation of GIN by the authors of the paper is \href{https://github.com/weihua916/powerful-gnns}{available on github }, using the GIN-0 version.
We use the recommended hyperparameters, i.e., five layers of GIN with two layers of MLP for each, a hidden size of 64, and summation as the graph- and node-pooling method. 
For the pre-training of the models, we set the number of epochs to 64, and we used the Adam optimizer with a learning rate of 0.01. Then, we train our model using an 80-20 train-test split on all our datasets.
We use the \href{https://github.com/clovaai/generative-evaluation-prdc}{PRDC library} by \cite{naeem2020reliable} and the Frechet distance implementation from pytorch-fid \cite{Seitzer2020FID}.
For the perturbation experiment, we use the absolute Spearman rank correlation coefficient (see \cite{zwillinger_crc_2000}) to assess the ranking ability of the given metric, with good ranking ability evaluated with an absolute value close to one and lousy ranking ability evaluated close to zero.

\ifthenelse{\boolean{article_extended}}{
\paragraph{Perturbation experiment}
In \autoref{tab:rank-perturb}, we observed that using a more extensive and more diverse training dataset impacts the behavior of FD positively and significantly, reducing the variance and improving the ranking ability. The F1-Score metrics exhibit more stability regarding the implementation except on the BA dataset and lower absolute Spearman rank correlation. We get a lower value for the F1-Score as they show high sensitivity towards such random perturbation on graphs, meaning that from a certain perturbation threshold, usually around 10 or 20\%, both F1-Score fall down to 0.
In \autoref{tab:rank-perturb-feature}, we observe that on the perturbation of grid graphs, the three given node feature initialization gave the same ranking coefficient, also with perfect stability through independent runs.
Now we review an experiment closer to a real graph generation model benchmark.
}{}

\paragraph{Grid experiment}
In \autoref{tab:GINmesSumFull}, we first observe that all metrics, computed using the degree node feature initialization, are consistently ranking the resampled grid graphs first when using the node, followed by GRAN generated grid graphs and the grid graph generated by both GraphRNN variants, which is consistent with the base assumption (see~\autoref{fig:grid}).
We also observe how The F1-Score metrics (PR and DC) show way less variance than the FD through independent runs. FD variance between independent runs is affected by the training dataset configuration, with the full dataset heavily reducing the variance for the resampled and GRAN-generated grid graphs. F1-Score is less affected by the training dataset configuration. The choice in the embedding function inside GIN (summation $x_{\sum}$ or concatenation $x_{\mathbin\Vert}$) does not change the behavior of all used metrics.
In \autoref{tab:GINmesranda}, we see that using other node feature initialization techniques, such as constant and random node features, gives inconsistent metrics results, with all tested graph embeddings-based metrics, in terms of ranking in the grid experiment.

\paragraph{Key Takeaways}
Like kernel-based measurement techniques, graph-embedding metrics behavior is highly sensitive to underlying hyperparameters \cite{obray2021evaluation}. More precisely, graph embeddings-based metrics that rely on trained graph classifiers are highly sensitive to more factors related to the training process, leading to reduced reproducibility and limited applicability of these metrics.
Manifold-based metrics in graph embedding space can alleviate the sensitivity issue, making the selection of such metrics more interesting. 
Graph embeddings-based metrics use is more exciting in benchmarking graph generators compared to kernel-based metrics as they can capture global information of the graph.
For instance, in the last section, two well-known graph generators were also compared using graph embeddings-based metrics.

\subsection{Comparing GRAN and GraphRNN}
We evaluate two state-of-the-art graph generation methods using the kernel-based MMD on distributions of graph invariants and previously explored metrics in graph embedding space.
For both models, we used the official github source code at \href{https://github.com/JiaxuanYou/graph-generation}{JiaxuanYou/graph-generation} and \href{https://github.com/lrjconan/GRAN}{lrjconan/GRAN} with recommended hyperparameters.

\paragraph{Experiment Parameters}
In this experiment comparing GRAN and GraphRNN, we also varied the used node ordering policy used to reduce the complexity graph adjacency matrices space to learn from.
For GRAN, we use BFS and DFS that is computed using the node with the highest degree, and other node ordering functions such as the default node ordering, degree-descent, and their denoted K-core ordering \cite{DBLP:journals/corr/abs-1910-00760}.
For GraphRNN, we use the vanilla BFS version, the uniform random ordering, our introduced DFS ordering, and a BFS variant where we do not reduce the graph matrix representation.

\paragraph{Experimental Setup}
For GraphRNN, we use 4-layer GRU cells with a 128-dimensional hidden state to encode the graph information and a 16-dimensional hidden state to encode the edge information.
The output of the edge-level RNN is mapped from 16 to 8 dimensions using an MLP, then passed through a ReLU to be mapped to a scalar to pass through a sigmoid activation function finally.
The model is trained on each dataset in a minibatch of size 32 using the Adam optimizer for 3000 epochs each, with a learning rate of 0.001 reduced by 0.3 at the 400th and 1000th epoch.
For GRAN, we keep the same hyperparameters for the model, using GNN with 7 layers and the block size and stride fixed at 1 because this will assure the highest generation quality possible.
The model also uses an Adam optimizer for the training phase and is trained with a learning rate of 0.0001 and no decay.
We train both models on the \textbf{BA}, \textbf{WS}, \textbf{C2L}, and \textbf{C2S} graph datasets using several node ordering, using 80-20 train-test splits. 

We compared GRAN and GraphRNN evaluation results and also added two reference measurements for each dataset in order to give a sense of ``scale'', as recommended by O'Bray et al. \cite{obray2021evaluation}. 
First, the evaluation compares the reference set of graphs with a resampled set of graphs from the same generator denoted with $\varheart$ or "Resampled," a very similar set of graphs.
Furthermore, we use the \textbf{ER} graph model as a benchmarking reference \cite{DBLP:journals/corr/abs-1910-00760}, computing metrics for a random graph generative model compared to the test graph sets.
We resample an instance of BA, WS, C2L, and C2S graph datasets using their base generation function, which will be compared to the generated graph datasets.
We train GIN on the full dataset configuration with one-hot node degree feature initialization and concatenation graph embedding function for Graph-Embeddings-based measurements. We then use the BA, WS, C2L, and C2S sets from the training dataset, which will be compared to generated graph sets.

\paragraph{Results}
Measurements based on kernel computation and those from GNN training give consistent comparisons between GraphRNN and GRAN (compare \ref{tab:com} and \ref{tab:GINmes}).
Through our experiments and measurements, we find that generally, GRAN generates graph lists that are more faithful to the base dataset, especially on medium-sized datasets like BA or WS in table \ref{tab:wsba}. 
However, GraphRNN shines when trained on small graph datasets like on C2S (compare \ref{tab:com}).
Across several tested node orderings for both GRAN and GraphRNN, we found discrepancies in generation quality, especially on GRAN. 
Our proposed DFS ordering for GraphRNN offers good performance, competing with the BFS ordering on our datasets, and even offers better performances on small datasets.

\section{Conclusion \& Future Work}\label{sec:conclusion}
This work compares GraphRNN and GRAN on different node orderings. 
We notice cases where GRAN wins, especially on medium-sized graphs, while GraphRNN has an advantage on smaller graphs.
The node ordering influences these models' performance, especially the generated graphs' quality.
Through extensive experiments, we have traced several insights into using GNNs to evaluate graph generators in expressiveness and stability.
The first insight is with changing pre-training datasets, where we show that augmenting pre-training datasets with Erdős-Rényi-graphs can improve the performance of the GNN for graph generator evaluation.
Also, we recommend using one-hot degree vectors to initialize the features of the nodes, which have shown the best results in terms of expressiveness and stability of the metrics.
Finally, we find that using manifold-based metrics also reduces the influence of other hyperparameters on the evaluation expressiveness and stability, in line with an observation made by \cite{thompson2022on}, thus giving our recommendation to favor these metrics as reference metrics for future studies of GGMs.
For future work, this work raises the interest in looking at other node orderings for using GraphRNN on small graphs.
Elaborate feature engineering approaches for graph nodes could improve the behavior of graph-embedding-based metrics.

\bibliographystyle{splncs04}
\bibliography{\contentBasePath bibliography}

\ifthenelse{\boolean{article_extended}}{
    \appendix
    \section{Appendix}
    \paragraph{Used node ordering for autoregressive models}
For both GraphRNN and GRAN, we trained those models independently using several node ordering which are describe as follows :
\begin{itemize}
    \item \textbf{BFS-Deg}: A BFS node ordering built by the BFS tree of each graph with the largest degree node as its root
    \item \textbf{DFS-Deg}: A DFS node ordering variant similar to BFS
    \item \textbf{Default}: The default node ordering used by NetworkX \cite{hagberg2008exploring} graphs
    \item \textbf{Degree Descent}: An ordering based of the degree of nodes (in descent order)
    \item \textbf{K-core}: A novel node ordering introduced by GRAN, called k-core ordering, built from the k-core graph decomposition
\end{itemize}

We compare the GraphRNN method under four different versions:
\begin{itemize}
    \item \textbf{BFS-Random}: A BFS node ordering built by the BFS tree of each graph where its root node is selected randomly, which is the default node ordering, which benefits from it by reducing the size of the graph adjacency matrix from $O(N^2)$ to $O(MN)$ 
    \item \textbf{DFS-Random} (Novel): We implemented a DFS node ordering variant similar to the default BFS one but without the benefits of using a reduced adjacency matrix, specific to BFS
    \item \textbf{BFS-Max}: Similar to BFS-Random, however, we used the complete graph adjacency matrix for DFS-Uniform. This will be useful to alleviate any bias linked to the choice of the hyperparameter M when comparing with DFS-Uniform
    \item \textbf{Default-Uniform}: The node ordering for each graph are completely shuffled at random
\end{itemize}

\paragraph{Additional experiments results}

\begin{figure}
    \centering
    \subfloat[Base Dataset]{\includegraphics[width = 0.33\textwidth]{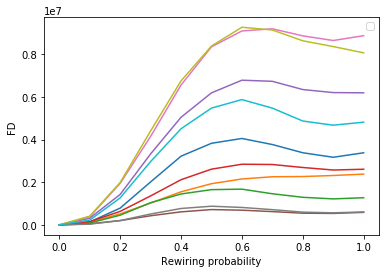}}
    \subfloat[Ladder Dataset]{\includegraphics[width = 0.33\textwidth]{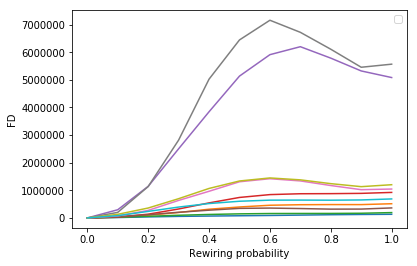}}
    \subfloat[Full Dataset]{\includegraphics[width = 0.33\textwidth]{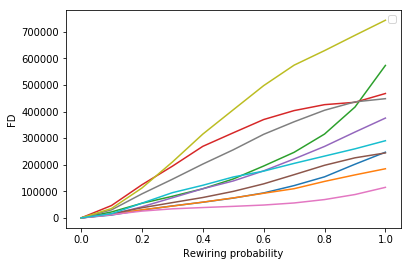}}\\
    \caption[FD metrics for Grid Graphs, using concatenated graph features]{
        Plots of Frechet Distance for perturbed Grid Graphs, over 10 indepedent trainings of GIN.
        Graph features are extracted with \textbf{concatenation}.
    }
    \label{FDCONCAT}
\end{figure}

\begin{figure}
    \centering
    \subfloat[Base Dataset]{\includegraphics[width = 0.33\textwidth]{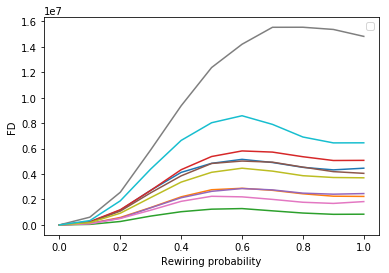}}
    \subfloat[Ladder Dataset]{\includegraphics[width = 0.33\textwidth]{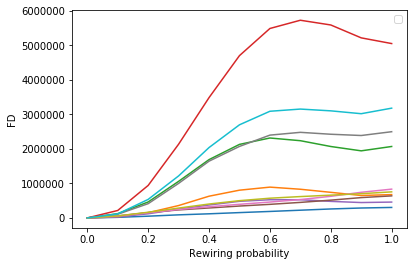}}
    \subfloat[Full Dataset]{\includegraphics[width = 0.33\textwidth]{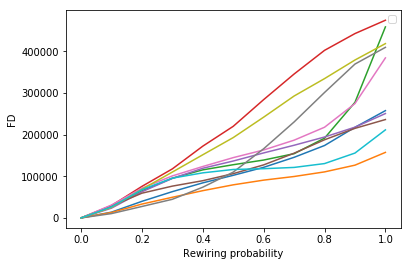}}\\
    \caption[FD metrics for Grid Graphs, using concatenated graph features]{
        Plots of Frechet Distance for perturbed Grid Graphs, over 10 indepedent trainings of GIN.
        Extracting graph features with \textbf{summation}.
    }
    \label{FDsum}
\end{figure}

\begin{figure}
    \centering
    \subfloat[Base Dataset]{\includegraphics[width = 0.33\textwidth]{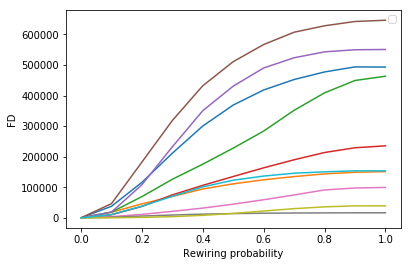}}
    \subfloat[Full Dataset]{\includegraphics[width = 0.33\textwidth]{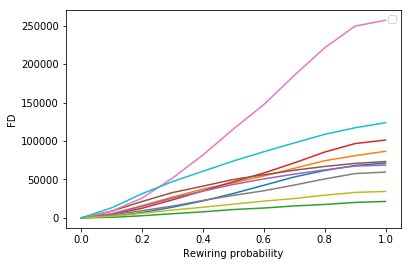}}\\
    \caption[FD metrics for Barabasi Graphs, using concatenated graph features]{
        Plots of FD for perturbed Barabasi-Albert Graph, over 10 indepedent trainings of GIN, extracting graph features with \textbf{concatenation}.
    }
    \label{FDconcatBA}
\end{figure}

\begin{figure}
    \centering
    \subfloat[Base Dataset]{\includegraphics[width = 0.33\textwidth]{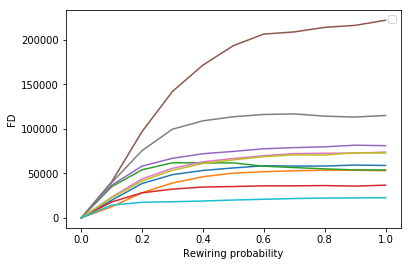}}
    \subfloat[Full Dataset]{\includegraphics[width = 0.33\textwidth]{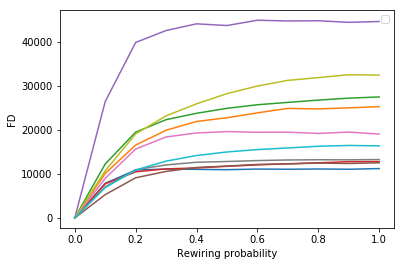}}\\
    \caption[FD metrics for Watts Graphs, using concatenated graph features]{
        Plots of FD for perturbed Watts-Strogatz Graph, over 10 indepedent trainings of GIN, extracting graph features with \textbf{concatenation}.
    }
    \label{FDconcatWS}
\end{figure}

}{}

\end{document}